\documentclass[letterpaper, 10 pt, conference]{IEEEtran}% Comment this line out if you need a4paper

\IEEEoverridecommandlockouts                              % This command is only needed if 
\usepackage{soul}
\usepackage[ruled, lined, linesnumbered,  longend]{algorithm2e}
\SetKwInput{KwInput}{Input}                % Set the Input
\SetKwInput{KwOutput}{Output}

\usepackage[dvipsnames,table,xcdraw]{xcolor}
\usepackage[utf8]{inputenc}
\usepackage{graphicx}

\usepackage{acro}
\usepackage{nomencl}
\makenomenclature

\usepackage{etoolbox}
\renewcommand\nomgroup[1]{%
  \item[\Large\bfseries
  \ifstrequal{#1}{N}{Nomenclature}{%
  \ifstrequal{#1}{A}{List of Abbreviations}{}}%
]\vspace{10pt}}
\begin{document}

\title{\LARGE \bf
Collision Detection: An Improved Deep Learning Approach Using SENet and ResNext
}

\author{\IEEEauthorblockN{Aloukik Aditya, Liudu Zhou, Hrishika Vachhani, Dhivya Chandrasekaran, Vijay Mago}\\
\IEEEauthorblockA{Department of Computer Science\\ Lakehead University\\
Thunder Bay, Ontario, Canada\\
Email: \{aaditya, lzhou9, hdhirend, dchandra, vmago\}@lakeheadu.ca}}

\maketitle
%\thispagestyle{empty}
%\pagestyle{empty}

%%%%%%%%%%%%%%%%%%%%%%%%%%%%%%%%%%%%%%%%%%%%%%%%%%%%%%%%%%%%%%%%%%%%%%%%%%%%%%%%
\begin{abstract}

In recent days, with increased population and traffic on roadways, vehicle collision is one of the leading causes of death worldwide. The automotive industry is motivated on developing techniques to use sensors and advancements in the field of computer vision to build collision detection and collision prevention systems to assist drivers. In this article, a deep-learning-based model comprising of ResNext architecture with SENet blocks is proposed. The performance of the model is compared to popular deep learning models like VGG16, VGG19, Resnet50, and stand-alone ResNext. The proposed model outperforms the existing baseline models achieving a ROC-AUC of $0.91$ using a significantly less proportion of the GTACrash synthetic data for training, thus reducing the computational overhead.

\end{abstract}

%%%%%%%%%%%%%%%%%%%%%%%%%%%%%%%%%%%%%%%%%%%%%%%%%%%%%%%%%%%%%%%%%%%%%%%%%%%%%%%%
\section{INTRODUCTION}

Vehicle collision is one of the significant causes of death in the transportation industry. Human errors and distractions are the leading causes of fatal accidents, and these deaths can be prevented using collision prediction techniques. The main aim of collision prediction systems (collision avoidance systems) is to predict the occurrence of collision as early as possible. In order to achieve this, a wide variety of input data is required, based on which the probability of collision is estimated \cite[]{taghvaeeyan2013two}. These include velocity, thrust, torque, video feeds, data from accelerometer sensors, gyroscopic sensors, and resting episodes \cite[]{ba2017crash}.

Each type of input data has its own significance and contributes uniquely to optimally predict peripheral detection, lane marking detection, and vehicle detection thus aiding in collision prediction. Hence identifying the correct data or combination of data is critical for effective prediction of collision. Recent research use sensor inputs from Light Detection and Ranging (LiDAR) systems and images from monochrome cameras to achieve better performance in collision avoidance systems. Advanced Driving Assistance System (ADAS) \cite[]{bayoudh2021transfer} and Automatic Emergency Braking (AEB) \cite[]{fildes2015effectiveness} systems use various combinations of input data to assist driving and automatic application of brakes when in risk-prone areas.

However, the addition of these sensors not only increases the cost of these vehicles but also poses challenges in their ergonomic design. To address this issue researchers have attempted to exploit the advancements in the field of computer vision. But deep learning models require significantly large data to effectively predict collision and collecting labeled data of real-world accidents is quite challenging. \cite[]{kim2019crash} built a hybrid dataset comprising videos of both real-world accidents captured and uploaded on YouTube\footnote{https://www.youtube.com/} and synthetic data generated using the video game Grand Theft Auto-5 (GTA). The dataset comprises of approximately $750,000$ samples of accident and non accident incidents. They further demonstrate the performance of various convolutional neural network models and the best performance (AUC $0.9097$) is achieved by the ResNet50 model.

In this article we use ResNext - an enhanced version of Resnet, appended with Sequence and Excitation (SE) blocks to achieve increased performance by exploiting approximately only 4\% of the training data, thus reducing the compuational overhead significantly. We further compare the performance of various models like VGG16, VGG19, ResNet50, Efficientnet\_B5, ResNext with our proposed model. Our model outperforms these models achieving a ROC-AUC of $0.9115$. There have been various attempts to develop an elegant solution for a collision-avoidance system, and researchers have exploited recent advancements in deep learning models to achieve better performance. But these models require large datasets and are computationally expensive. The motivation behind this research is to offer a solution that outperforms the existing models while reducing the computational overhead by utilizing lesser number of data points. \\
\indent The remainder of the article is organized as follows. Section \ref{related} discusses the related work carried out in the field of Automated Collision Detection. Section \ref{method} describes in detail the methodology of the proposed model. Section \ref{results} presents results obtained and Section \ref{conclusion} concludes the work by summarizing the work done providing the scope of future research.

\section{Related work}\label{related}
\subsection{Collision prediction algorithms}
For an efficient collision detection system, predicting the trajectory and motion of the vehicle is crucial. The survey by \cite{lefevre2014survey} discusses the different approaches that can predict a vehicle's trajectory. It compares the trade-off between the model completeness and real-time implementations of such systems. The survey also highlights the fact that the choice of risk assessment methods is dependent on motion prediction models. They have categorized the motion prediction models into three main categories. First, the physics-based motion model - where the model depends on the law of physics; second, the maneuver-based models - where the model considers the maneuvers that the driver intends to perform. And finally, the interaction-aware motion models - where models consider the inter-dependencies between different vehicles' maneuvers.

Awareness of the surroundings of the vehicle can prevent accidents that occur between the vehicle and peripheral objects such as pedestrians. For peripheral detection systems, \cite{ahmad2017symbolic} have proposed a method using five different architectures of CNN networks to recognize the painted symbolic road markings (lane markings). Their non-conventional approach, which consists of MSER (Maximally Stable Extremal Regions) feature extraction of the IPM (Inverse Perspective Mapping) image, has proven advantageous.

The idea of using deep reinforcement learning in which algorithms learn from the experience of trial and error has been proven to be useful in the navigation system. For instance, the collision avoidance mechanism by self-supervised learning using a standard stereo camera with the navigation model outperforms the Double Q-learning method resulting in a fully autonomous navigation experience \cite[]{kahn2018self}. Double Q-learning is a part of deep reinforcement leaning and their approach has seventeen times farther crash distance than random policy and seven times more distant than the double Q-Learning method. Deep reinforcement learning can also be advantageous in decentralized vehicular networks. \cite{chen2017decentralized} have implemented a application of deep reinforcement learning, which uses a decentralized collision avoidance algorithm for finding the optimal path. They have developed a value network that encodes the destination's expected time considering neighbors' position and velocity. Based on the particular vehicle's value network and surrounding vehicle's value network, the optimal path with the lowest collision likelihood is predicted. The simulation for multiple vehicles having source and destination combination resulted in $26$ \% improvement in the optimal path's quality than the general path without considering other vehicles. In another approach, \cite{long2017towards} have taken a step further by using deep reinforcement learning in decentralized vehicular collision avoidance mechanism. Their approach has used sensors to communicate with their Policy Gradient Algorithm, which consists of parameters such as success rate, collision avoidance performance, and generalization capability. This method consists of numerous rich and complex environments resulting in time-efficient and collision-free paths in a massive scale decentralized system.

Intersections on the roads are the weakest point because the driver needs to focus on various aspects like the traffic signal, the incoming traffic, the pedestrians, etc. Due to this, collisions at the intersections are prominent. \cite{kim2020unexpected} designed intelligent self-driving policies that reduce the severity of the incidents at the crossroads, majorly due to traffic signal violations. They used a deep reinforcement learning approach that provides collision avoidance irrespective of the intersection's destination end. This approach outperforms both human drivers and Autonomous Emergency Braking (AEB) system. However, as the authors mentioned, there are some limitations, for instance, the inability to detect lanes on the road because of only lidar sensors.

Collisions at the back of the vehicle are frequent, as the driver can not continuously monitor the vehicle's back end. \cite{chen2018rear} proposed a method based on a GA-optimized Neural Network that calculates the variables and controls a collision's likelihood at the rear end of the vehicle. They also used deep learning using BP (Back Propagation) model to optimize collision risk parameters. As a result, they achieved a Mean Square Error (MSE) of nearly zero.

Detecting traffic accidents in major cities is even more challenging than in rural areas due to more vehicles, roads, and intersections. \cite{parsa2020toward} used the Chicago dataset\footnote{Dataset collected and archived by the Illinois Department of Transportation (IDOT). The accident cases occurred on the Chicago metropolitan expressways between December $2016$ and December $2017$} which contains $244$ accident cases and $6,073$ non-accident cases which occurred on Chicago metropolitan expressways. They used the Extreme Gradient Boosting (XGBoost) technique to predict accidents in real-time. They have considered various factors such as traffic, demographics, weather, and many more for prediction. In addition to that, the authors have done feature dependency analysis on the data and achieved a significant increase in the real-time accident detection rate.

Numerous machine learning methods have been applied to self-driving systems resulting in better detection systems. \cite{badue2020self} have incorporated a complete overview of self-driving car's system architecture, organized into perception system and decision-making system of vehicles. They have tokenized their thorough research of perception systems based on self-driving-car localization, moving obstacles tracking, static obstacles mapping, and road mapping. Decision-making systems help the vehicle navigate from starting location to the final spot; it is divided into several subsystems, such as route planning, motion planning, control, and behavior selection.

\subsection{Domain adaptation} Collecting and labeling quality datasets is both costly and time-consuming. One of the solutions to address this problem is generating synthetic data. However, it is essential to understand that there is a significant gap between real and synthetic data. In order to bridge this gap, domain adaptation techniques are employed to annotate unlabelled datasets using a labeled dataset from the same domain.

One approach to solving the domain gap between real and synthetic data is to use both synthetic data and real data when training a task. \cite{richter2016playing} presented an approach to produce pixel-accurate semantic label maps for images synthesized by modern computer games and have shown that this approach can increase the performance of semantic segmentation models on real-world images. This idea can be extended to produce a continuous video stream. \cite{wang2019learning} is the first to present a crowd counting method via domain adaptation; by realizing this method, the domain gap between the synthetic and real data can be significantly reduced.

Another approach is unsupervised domain adaptation, which can solve a significant performance drop when a model is trained on a source domain and tested on a target domain. Particularly, in image level adaptation, images collected from different cameras such as DSLR cameras and web cameras are much different. A classifier trained on the DSLR cameras may fail to train on the web cameras without adaptation. To build an efficient classifier, it is necessary to consider the shift between these two distributions. \cite{hoffman2018cycada} at first used a Generative adversarial Network(GAN) to transfer the image styles from the source domain to a target domain, so the transferred image has a similar style with the target. Second, the style-transferred images and the associate labels were used in supervised learning in the target domain. Generally, the aim is to train powerful classifiers for the target samples; it is known that a GAN objective function can be used to learn target features identical from the source ones. \cite{volpi2018adversarial}, extend this GAN framework by enforcing domain invariance and performing feature augmentation which leads to superior performance.

With the successful development of GANs, image-to-image translation tasks are dominated by GANs because of the high fidelity in generated images. \cite{isola2017image} have used conditional GANs to propose a model known as Pix2Pix where they condition on an input image to generate a corresponding output image. They use their model to test various tasks and datasets, including graphics tasks (photo generation) and vision tasks (semantic segmentation.) The conditional GANs map from an input image to an output image and learns loss function to train this mapping. \cite{dong2017unsupervised} propose a two-step learning method that utilizes conditional GAN to learn the global feature of different domains using unlabeled images without specifying any correspondence between them; it also helps in avoiding the need for labeled data.

Obtaining paired (two similar image datasets) training data can be difficult and expensive; \cite{zhu2017unpaired} have shown success dealing with unpaired data. They presented a system called CycleGAN, which learns to capture unique characteristics of one image collection and apply these features to a new image. Their method often succeeds in translation tasks involving color and texture changes but with little success when exploring tasks requiring geometric changes.

Unsupervised image-to-image translation attempts to learn a mapping that can translate an input image into a similar image in two different domains. Even though the existing unsupervised image-to-image translation models can generate realistic translations, they still have some drawbacks, like the need for extensive training datasets and the inability to create images from unseen domains. \cite{liu2019few}, argue the limits they use and seek a few-shot about the unsupervised image to image translation framework. Their model achieves this few shot generation capability by coupling an adversarial training scheme with a novel network design. They verify their effectiveness of the proposed framework by the comparisons of several baseline methods on benchmark datasets.

Some works extend the unsupervised image translation to multiple domains. They learn a mapping between various seen domains simultaneously. \cite{choi2020stargan} show improved results over the following properties in an excellent image to image translation model; diversity of generated images and scalability over multiple domains. To address scalability, they propose a single framework called StarGAN v$2$ which is an extensible approach that can generate diverse images across multiple domains. However, An earlier model (StarGAN) learns the mappings between all available domains using a single generator. StarGAN v$2$ can generate images with rich styles across numerous domains, outperforming the former methods.

\section{Methodology}\label{method}
\subsection{Dataset}

\begin{figure}[htp]
    \centering
    \includegraphics[width=9cm]{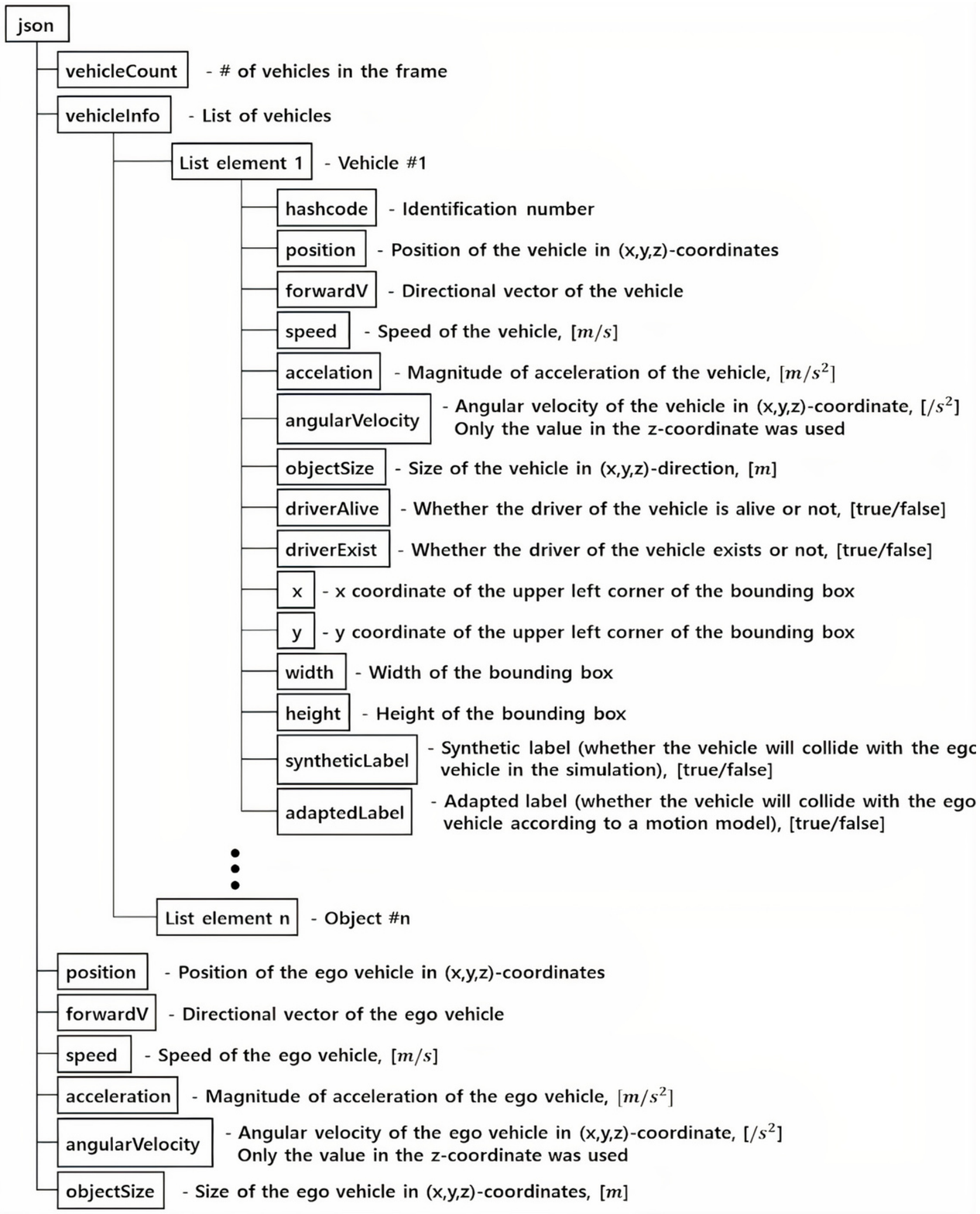}
    \caption{GTACrash JSON file structure}
    \label{fig 1}
\end{figure}

The proposed model is trained using synthetic data and tested on real-world data. The process adopted by \cite{kim2019crash} to build the GTACrash and YouTubeCrash datasets is described in this section. Both datasets have two classes, $accident$ and $non-accident$, positive cases are labeled as accident, negative cases are labeled as non-accident. Each class has JPEG and JSON files. The raw JPEG image resolution is $400 \times 710$ pixels, but it was reduced to $130 \times 355$ pixels during the preprocessing. The JSON files have sensitive information about the vehicle scenario, including coordinates, wheel angle, acceleration, angular velocity, etc. as shown in Figure \ref{fig 1}.

The game GTA-5 was played for $72$ hours straight using the automated script of the GTACrash synthetic data in the original paper, which produced $3,661$ non-accident scenes and $7,720$ accident cases. Each case has $20$ frames and $20$ JSON files, making the total accident frame $20,000$ and non-accident to $10,000$. The images were collected under various day and weather conditions like sunny, rainy, nights, snow, etc. These attributes positively affect the algorithm of this project. The software called ScriptHook V was used to attain the data; one of the attributes was bounding\_box, it extracted the coordinates of the X and Y axis for the nearby vehicles. These coordinates were later used by the algorithm to classify dangerous (red box) or safe vehicles (green box). However, we have reduced the original dataset and only use $1,000$ accidental cases and $500$ non-accident cases.

As shown in Table \ref{our-dataset}, YouTubeCrash data is split randomly into training and testing. This dataset was collected from dashcam videos (uploaded on a YouTube channel called Car Crashes Time). The videos in the YouTubeCrash dataset have $122$ clips which were further divided into accident and non-accident classes with 20 frames per case. Accident and non-accident images were collected using each dashcam video: if an accident happened at the $5$th second of the video, then from $3$ to $5$ seconds (just $2$ seconds before an accident), $20$ frames were collected and labeled as ``accident". While $0$ to $3$ seconds of the video was marked as ``non-accident".
The dataset contains diverse accident scenarios such as abrupt lane changes, sudden stops, and signal violations; it can help the model to solve real-world problems in future mixed conditions. There are JSON files for the YouTubeCrash dataset as well, but the parameters such as position, acceleration, and angular velocity are set to $NA$. As shown in Table \ref{our-dataset}, we have reduced the number of instances for the GTACrash dataset, but instances in the YouTubeCrash dataset were used the same as the original paper. Table \ref{original-dataset} illustrates the cases for the original dataset, while Table \ref{our-dataset} shows the instances of ours.\par

We also explore other car crash image datasets, including the KITTI dataset. \cite[]{zhu2018vision}, which were created in the city of Karlsruhe (urban as well as rural areas). In each image, there are up to 15 cars and 30 people (depends on the location), and the engine hood and the sky region have been cropped to achieve better performance. Also, both color and grayscale images were stored with lossless compression using $8$-bit PNG files. The second dataset is $Cityscapes$ \cite[]{Cordts2016Cityscapes}; it contains a diverse set of video sequences recorded in street scenes from 50 different cities, with high-quality pixel-level annotations of $5,000$ frames in addition to a more extensive collection of $20,000$ weakly annotated frames. These datasets helped us understand the diverse conditions of the accidents and non-accident scenes that occur in urban and rural areas but this research focuses on the use of synthetic data for training. 

\begin{table}[]

% increase table row spacing, adjust to taste
%\renewcommand{\arraystretch}{1.3}
% if using array.sty, it might be a good idea to tweak the value of
%\extrarowheight as needed to properly center the text within the cells
\caption{Statistics of Original YouTubeCrash and GTACrash}
\label{original-dataset}
\centering
% Some packages, such as MDW tools, offer better commands for making tables
% than the plain LaTeX2e tabular which is used here.

% Please add the following required packages to your document preamble:
% \usepackage[table,xcdraw]{xcolor}
% If you use beamer only pass "xcolor=table" option, i.e. \documentclass[xcolor=table]{beamer}

% Please add the following required packages to your document preamble:
% \usepackage[table,xcdraw]{xcolor}
% If you use beamer only pass "xcolor=table" option, i.e. \documentclass[xcolor=table]{beamer}
% Please add the following required packages to your document preamble:
% \usepackage[table,xcdraw]{xcolor}
% If you use beamer only pass "xcolor=table" option, i.e. \documentclass[xcolor=table]{beamer}

\begin{tabular}{|c|c|c|}
\hline
 
\textit{Number of} & {\color[HTML]{CB0000} \textit{YouTubeCrash}} & {\color[HTML]{3166FF} \textit{GTACrash}} \\ \hline
Orignal accident scenes    & 122  & 7720   \\ \hline
Orignal nonaccident scenes & 100  & 3661 \\ \hline
Total accident frames      & 2440 & 154400 \\ \hline
Total non accident frames  & 2000 & 73220  \\ \hline
\end{tabular}

\end{table}

\begin{table}
% increase table row spacing, adjust to taste
%\renewcommand{\arraystretch}{1.3}
% if using array.sty, it might be a good idea to tweak the value of
%\extrarowheight as needed to properly center the text within the cells
\caption{Statistics of Ours YouTubeCrash and GTACrash}
\label{our-dataset}
\centering
% Some packages, such as MDW tools, offer better commands for making tables
% than the plain LaTeX2e tabular which is used here.
\begin{tabular}{|c|c|c|}
\hline
\textit{Number of} & {\color[HTML]{CB0000} \textit{YouTubeCrash}} & {\color[HTML]{3166FF} \textit{GTACrash}} \\ \hline
Ours accident scenes      & 122  & 1000  \\ \hline
Ours nonaccident scenes   & 100  & 500   \\ \hline
Total accident frames     & 2440 & 20000 \\ \hline
Total non accident frames & 2000 & 10000 \\ \hline
\end{tabular}
\end{table}

% \subsection{Data handling and Preprocessing (Not ready, just blueprint for next week)}

% \begin{itemize}
%   \item File size reduced to 130X355.
%   \item Data is divided into chunks to avoid memory overload.
%   \item JSON and JPG files are combine to create a complete frame list.
%   \item To get vehicle labels Bounding box and other JSON inputs are used.
%   \item After getting frames and labels, data is divided into training, testing and validation.
%   \item Training is using gtat\_rain and YouTube\_train however testing is done using only YouTube\_test.
%   \item To increase performance , multi threading have been used with 16 thread for each instance.
%   \item Batches were created using FIFO queue and dequeue with the size of 512.
%   \item Imageaugmentationa and augmented\_box used to increase the perfomance.
  
% \end{itemize}

\subsection{Experimental Setup}

We have implemented and tested various models such as VGG16, VGG19, ResNext, Efficientnet B($0$-$7$), and ResNext\_SENet on the dataset and obtained different outcomes. %and Figure 4 shows the result for the models which we ran using our altered dataset; it can be seen that
Our proposed method, ResNext + SENet with ROC-AUC of $0.9115$, surpassed the existing methods Resnet50 and VGG16 with ROC-AUC of $0.9069$. We have also implemented other deep learning models but the best performing in terms of ROC-AUC and complexity is ResNext+SENet model.  %\textcolor{green}{Although, the implementation of all the models was done using TensorFlow version 1, they had minimal support for that version. \textcolor{blue}{Please decide what you are trying to tell here. Makes no sense.}}
%have implemented all our models using TensorFlow 1; many of our models had minimal support for TensorFlow 1.\par

\begin{algorithm}[h]
\DontPrintSemicolon
\SetAlgoLined
\SetNoFillComment
\LinesNumbered
%Algorithm 1 
\KwInput{Images and JSON file as $D_{whole}$ ($D_{train}$, $D_{test}$, $D_{val}$)}
\KwOutput{$dequeue$}

$reduce\_size$ $\gets$ reduce\_size($D_{whole}$)\\ \tcp*{resize to 130 x 355} 

$JSON\_files \gets$ build\_sample\_list($D_{whole}$) \\ \tcp*{reading JSON parameter: bounding\_box, wheel\_angle etc.}
$merge \gets$ frame\_per\_vehicle($reduce\_size ,JSON\_files$ )\\ \tcp*{JSON and img merged as one}

$preprocessing \gets$ preprocess\_batch($merge$)\\ \tcp*{data augmentation, resize boundingbox}

$enqueue \gets$ enqueue\_batch($preprocessing$)\\ \tcp*{batch of 512 is collected}

$dequeue \gets$ dequeue\_batch($enqueue$)\\ \tcp*{batch of 512 is ready to release into neural network}
\Return $dequeue$
\caption{Pseudocode for Preprocessing of Data}
\label{algo: AlgorithmPre}
\end{algorithm}

% 1) Dataset
% 2) Experimental Setup
% 3) Computation Resources
% 4) Results and Discussion

As shown in Algorithm \ref{algo: AlgorithmPre}, there are training, testing, and validation sets, and each set has an equal number of images and JSON files selected from the respective folders in the dataset. Initially, the image size is reduced from $400 \times 710$ pixels to $130 \times 355$ pixels and then read the JSON file parameters such as acceleration, wheel angle, angular velocity, etc. Using the merge function, the images and JSON files are merged as one instance. After that, preprocessing has been done using data augmentation and bounding\_box resize, which also helped to get the bounding box coordinates. Finally, using $enqueue$ and $dequeue$ functions, the input data has been passed into the neural network with a batch size of $512$, which helped avoid memory errors.

% Intro to ResNext
\subsubsection{ResNext}
\begin{figure}[htp]
    \centering
    \includegraphics[width=8cm]{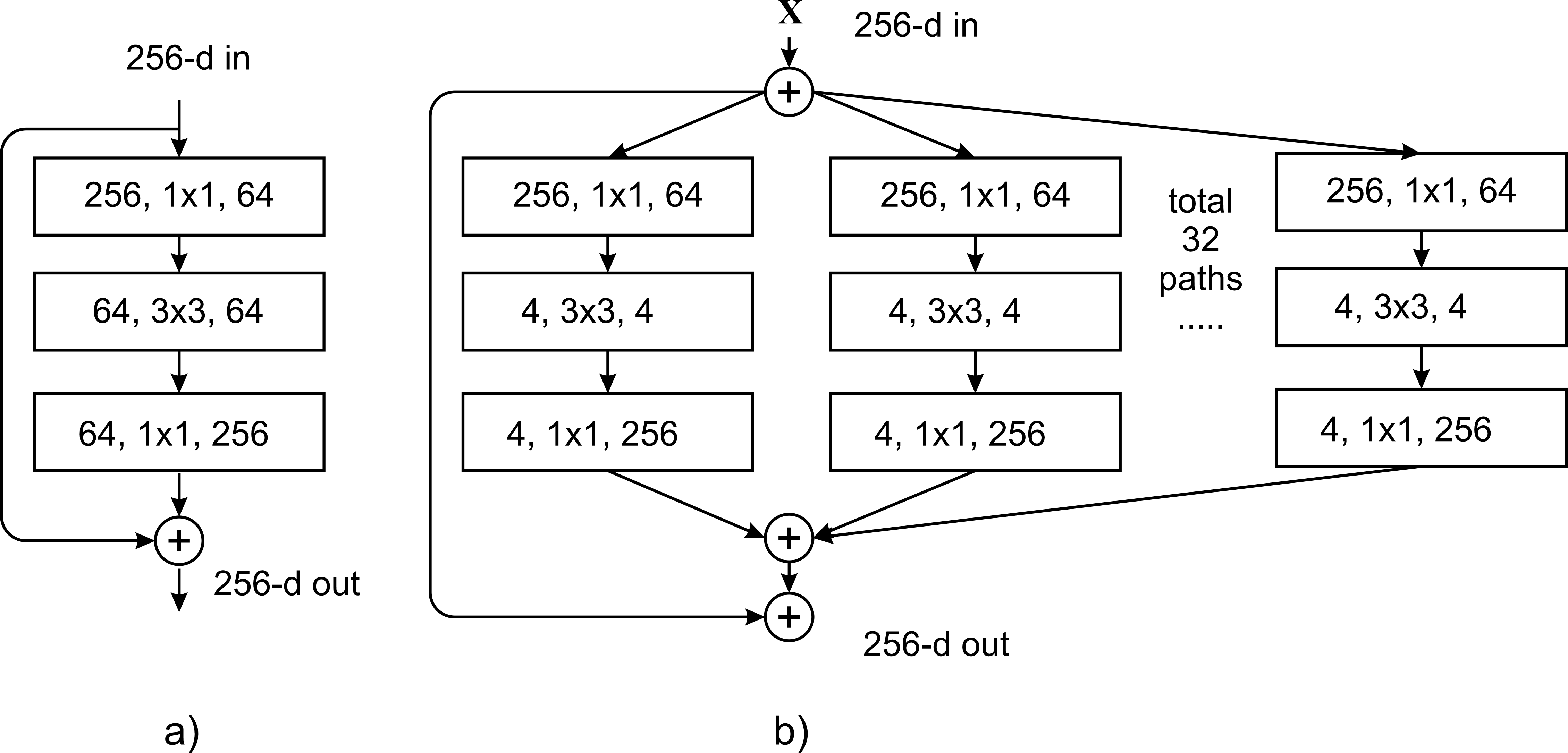}
    \caption{Resnet vs ResNext from \cite{8100117}}
    \label{fig 2}
\end{figure}

ResNext is an enhanced version of Resnet, which has one extra dimension called cardinality. This new dimension is the basically the size of the set of transformations \cite[]{8100117}. Cardinality splits the network into $n$ branches, and these branches perform convolution transformations that aggregate together using summation, as shown in Figure \ref{fig 2}. The left side is a block of Resnet, and the right side indicates ResNext with a cardinality of $32$. Also, Resnet is a particular case of ResNext in which cardinality is set to $1$ as there is only one path. ResNext is capable of performing better than Resnet but has 50\% more complexity. According to \cite{8578814}, ResNext can not perform well with small networks since it uses costly dense $1 \times 1$ convolution.

\subsubsection{Squeeze and excitation networks}

Squeeze and Excitation Networks (SENet) are the special types of networks that can be applied to any CNN network; they can improve channel inter-dependencies without computational cost \cite[]{hu2018squeeze}. In simple words, it can excite relevant features in the CNN layer and suppress irrelevant features without any additional processing. The main objective of convolution layers is to detect elements like edges, color, corners, etc. As we go deeper into the network, it can also detect shapes, digits, faces, etc., and all these features are fused with channel information of the image ($W \times H \times C$). SENet is generally applied to each convolution layer; it works best when the network is deep because of more spatial features. The weights of CNN channel are equal when a feature map is created, and these weights are modified after passing through SENet.

\begin{figure}[htp]
    \centering
    \includegraphics[width=8cm]{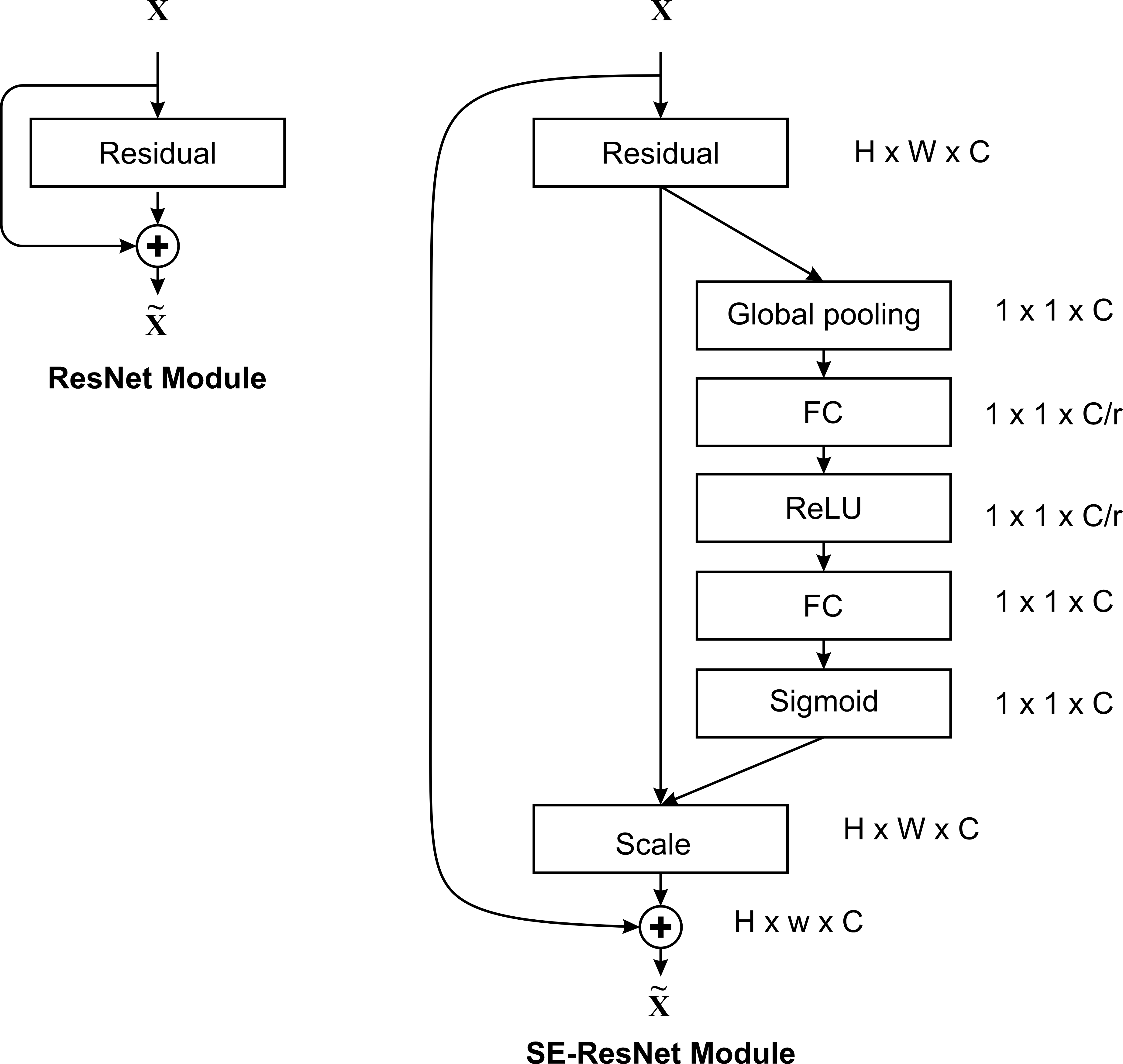}
    \caption{SENet architecture from \cite{hu2018squeeze}}
    \label{fig 3}
\end{figure}

The work of SENet is straightforward; it is a four-step network used in each convolution layer in the CNN network. Initially, we have an output from the convolution block ($W \times H \times C$), which is squeezed to $1 \times 1 \times C$ using global average pooling. Next, a fully connected layer followed by the Relu activation function is applied; it helps in getting the non-linearity to the features by removing all negative values. A fully connected layer followed by a Sigmoid activation function is implemented after that, resulting in shape $1 \times 1 \times C$. This final shape $1 \times 1 \times C$ is multiplied with the original input feature map of size ($H \times W \times C$), which basically scales the original input based on the important spatial features. The main task of the Squeeze and Excitation Network is to get the most critical features (excited by multiplying the vector in the final step) and to reduce irrelevant features (squeezed by global average pooing in the beginning).

\begin{figure*}
  \includegraphics[width=\textwidth]{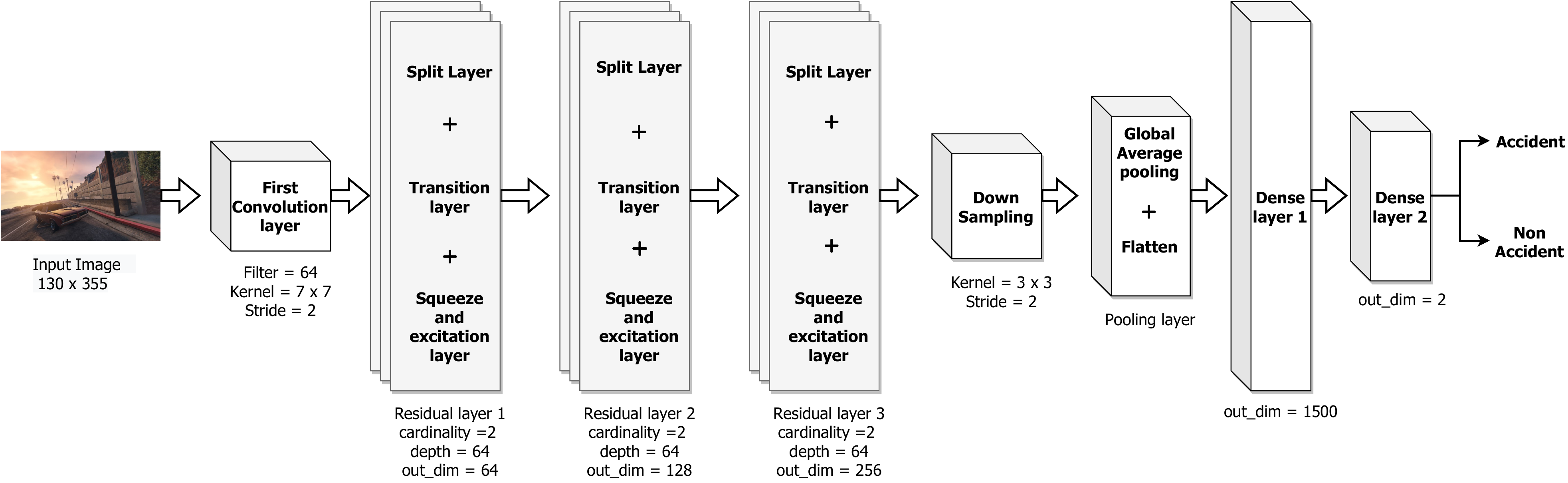}
  \caption{Architecture of ResNext + Squeeze and excitation network}
  \label{fig 4}
\end{figure*}

\subsection{Proposed Method}

The complete architecture of the proposed method can be seen in Figure \ref{fig 4}. In this research, a deep-learning-based model comprising of ResNext architecture with SENet blocks is proposed. Initially, normalization of the data using mean and standard deviation has been done, which helped remove the outliers or extreme values. Three residual layers have been used because of the bottleneck problem and over-fitting, and each of them contains three sub-layers: split, transition, and Squeeze-Excitation layer. Firstly, the split layer divides the layers into two branches as the cardinality is $2$. Secondly, the transition layer is the pooling layer plus the convolution layer, which results in the reduced dimension of the images. Finally, as discussed before, Figure \ref{fig 3} shows the Squeeze-excitation network, which helped achieve better ROC-AUC without any additional computational cost and processing time. After that, we passed the features to the generator layer, which downsampled the features, and by using global average pooling, it fixed the problem of over-fitting. Two dense layers were used as classification layers with a size of $1,500$ and $2$, respectively. Our model has a depth of $64$, which speeds up the processing time and makes it less complex. Also, in residual layers, we set the output dimension to $64$ in the first layer, and then it was increased to $128$ in the second layer and $256$ in the third layer.\par

 VGG16 and Resnet50 have been deployed and found that VGG16 not only has a reduced ROC-AUC but has a significant number of parameters. In addition to that, the training time of VGG16 was comparatively more than the Resnet50. The ResNext model gives the equivalent ROC-AUC compared to Resnet50, but the compilation time is relatively minor. A small additional method SENet added to the existing ResNext model, generates the best ROC-AUC, see section \ref{results}.\par
 
Adam optimizer is used in the proposed model because it can handle sparse gradient and noisy problems. Adam optimizer is not only adaptive, but it is highly correlated to Adagrad and Rmsprop optimizer. Our study implemented Adam, momentum, SGD, and Rmspropo optimizer; Adam optimizer performed best because our images deal with the vehicle motion.\par

\subsection{Computational resources}

Our project was supported by Compute Canada(http://www.computecanada.ca). We used the {\it Cedar} cluster, running on 24 CPU cores with 100GB memory and allocated time for 24.5 hours for eash task (the time for each task was about 4 hours with the output stored in a slurm file). The jobs were requested to run on 32GV100s.

\section{Results}\label{results}

While implementing ResNext, the variety of depth and width of the network have been experimented and observed that cardinality of $2$ with a depth of $64$ were the most optimized parameters for the network. 
%[rewrite]However, sometimes the model was not performing well due to high cardinality or width. 
Global average pooling, max pooling, and $tf.reduced\_mean$ were deployed and found that Global average pooling worked best in ResNext\_SENet because of synthetic data. However, in Resnet50 and VGG16, $tf.reduced\_mean$ works better than global average pooling because it computes the mean of elements across dimensions of a tensor. At the end of the structure of ResNext\_SENet, two fully connected layers ($1,500$ and $2$ neurons respectively) were used, which helped the model to understand the spatial features more accurately and resulted in the best ROC-AUC of $0.9115$. The hyperparameters of ResNext\_SENet were optimzed as shown in Figure \ref{Hyperparameters} which helped us acheiving better ROC-AUC from 0.81 to 0.9115.\par

% \usepackage[table,xcdraw]{xcolor}
% If you use beamer only pass "xcolor=table" option, i.e. \documentclass[xcolor=table]{beamer}
% \usepackage[normalem]{ulem}
% \useunder{\uline}{\ul}{}
\begin{table}[]
\caption{Hyperparameter tuning}

\centering
% Some package

\begin{tabular}{|c|c|c|}
\hline
Hyperparameters                           & {\color[HTML]{CB0000} Before} & {\color[HTML]{3166FF} Optimized} \\ \hline
Optimizer                                 & SGD                           & Adam                             \\ \hline
Cardinality                               & 8                             & 2                                \\ \hline
Depth                                     & 32                            & 64                               \\ \hline
Pooling                                   & tf.reduce\_mean               & Global\_Average\_pooling         \\ \hline
\multicolumn{1}{|l|}{Dense layer neurons} & 64                            & 1500                             \\ \hline
{\ul \textbf{ROC\_AUC}}                   & {\ul \textbf{0.81}}           & {\ul \textbf{0.9115}}            \\ \hline
\end{tabular}
\label{Hyperparameters}
\end{table}

EfficientNets are the most recent successful deep learning approach, but in this research, it was observed that results were not good as ResNext\_SENet. It is based on three parameters: width, depth, and resolution of the image. Here, width is the number of channels used in each layer, depth is the number of layers used in a network, and resolution is the the size of an input image. EfficientNet has its functionality called compound scaling; it can scale these parameters to get the most optimal and best accuracies with lesser flops and lesser parameters. Depth, width, and resolution are tuned using EfficientNet itself. All models of EfficientNet (B$0$-B$7$) were deployed, and each model has a different architecture based on image size. It was observed that EfficientNet B$5$ performed best with ROC-AUC of $0.85$ because it has the most similar image size architecture compared to synthetic images, which were used in this research. The performance of the EfficientNets could be improved by optimizing the hyperparameters but it requires significant amount of memory. \par

\begin{figure}[htp]
    \centering
    \includegraphics[width=9cm]{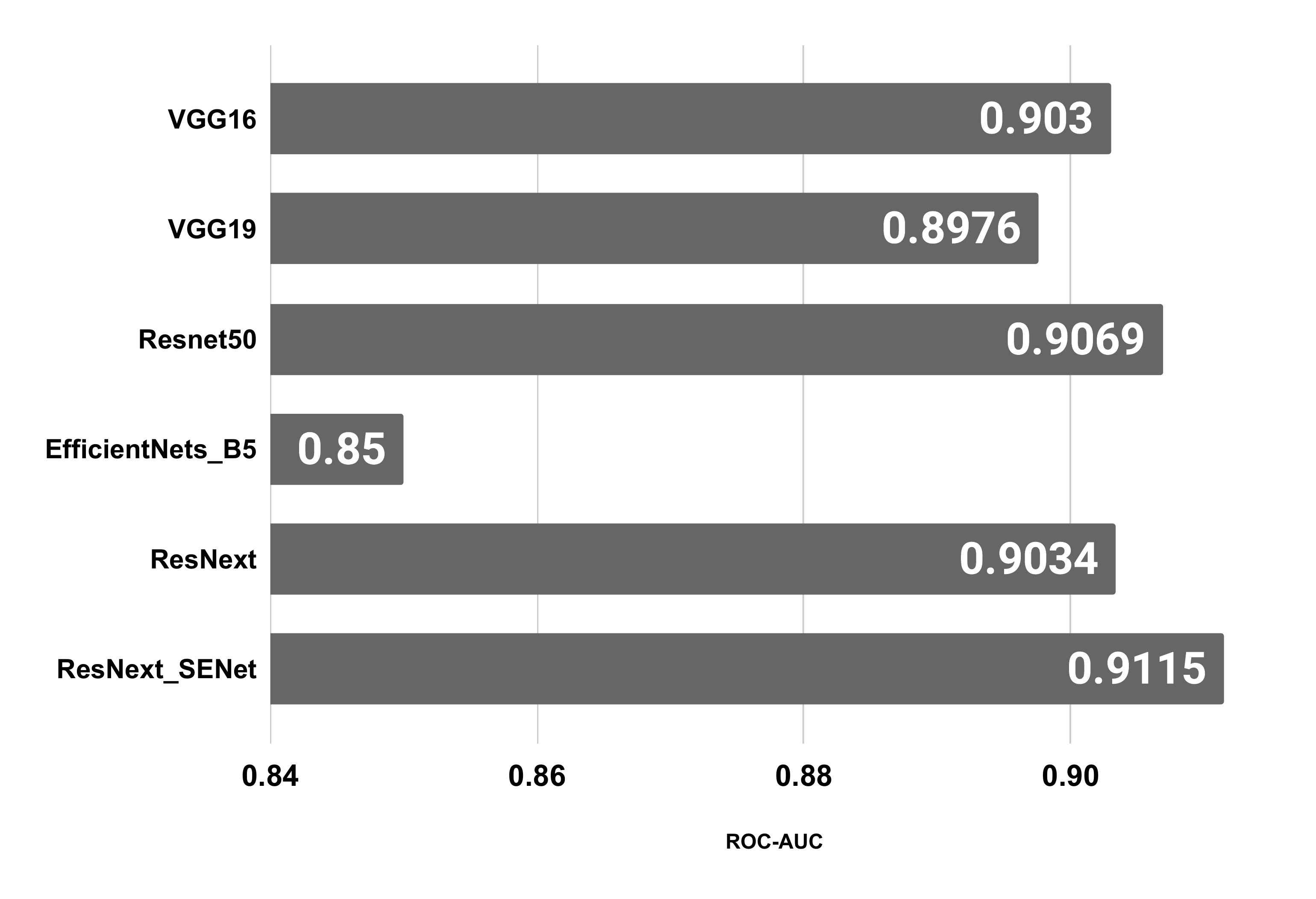}
    \caption{ROC-AUC performance comparison}
    \label{Result}
\end{figure}

ResNext\_SENet was used with sliced GTACrash dataset and achieved the ROC-AUC of $0.915$ as we can see in Figure \ref{Result}. Previously, \cite{kim2019crash} implemented Resnet50 with original GTACrash dataset which resulted the ROC-AUC of $0.9097$. In the future, we will deploy Rand augmentation and auto augmentation, which we believe can significantly improve the performance of the proposed model. \par

\section{Conclusion}\label{conclusion}

With the increasing number of deaths due to road accidents, research focus on building collision avoidance and collision prediction systems have gained significance. Researchers use data from various sources to efficiently predict accidents and provide assisstance to In this article a deep learning-based approach trained using synthetic data provided by \cite{kim2019crash} is proposed. The proposed model uses ResNext architecture with SENet blocks to outperform the existing base-line models while using significantly less training data. 

%\addtolength{\textheight}{-12cm}   % This command serves to balance the column lengths
                                  % on the last page of the document manually. It shortens
                                  % the textheight of the last page by a suitable amount.
                                  % This command does not take effect until the next page
                                  % so it should come on the page before the last. Make
                                  % sure that you do not shorten the textheight too much.

%%%%%%%%%%%%%%%%%%%%%%%%%%%%%%%%%%%%%%%%%%%%%%%%%%%%%%%%%%%%%%%%%%%%%%%%%%%%%%%%

%%%%%%%%%%%%%%%%%%%%%%%%%%%%%%%%%%%%%%%%%%%%%%%%%%%%%%%%%%%%%%%%%%%%%%%%%%%%%%%%

%%%%%%%%%%%%%%%%%%%%%%%%%%%%%%%%%%%%%%%%%%%%%%%%%%%%%%%%%%%%%%%

\section{Acknowledgments}

This research was enabled in part by support provided by Lakehead University (https://www.lakeheadu.ca/) and Compute Canada (www.computecanada.ca). 
The authors would like to thank Andrew Fisher for his support in implementation and development of the article. Publication costs are funded by NSERC Discovery Grant (RGPIN-2017-05377), held by  Vijay Mago.

\bibliographystyle{IEEEtran}
\bibliography{refs}

\vspace{-0.45cm}
\nomenclature[A]{{ADAS}}{Advanced Driving Assistance System}
\nomenclature[A]{{AEB}}{Automatic Emergency Braking}
\nomenclature[A]{{SENet}}{Squeezing and Excitation Network}
\nomenclature[A]{{ROC}}{Receiver Operator Characteristic}
\nomenclature[A]{{AUC}}{Area Under the Curve}
\nomenclature[A]{{MSER}}{Maximally Stable Extremal Regions}
\nomenclature[A]{{IPM}}{Inverse Perspective Mapping}
\nomenclature[A]{{BP}}{Back Propagation}
\nomenclature[A]{{MSE}}{Mean Square Error}
\nomenclature[A]{{XGBoost}}{Extreme Gradient Boosting}
\nomenclature[A]{{DSLR}}{Digital Single-Lens Reflex}
\nomenclature[A]{{GAN}}{Generative Adversarial Network}
\nomenclature[A]{{JSON}}{JavaScript Object Notation}
\nomenclature[A]{{JPEG}}{Joint Photographic Experts Group}
\nomenclature[A]{{CNN}}{Convolutional Neural Network}
\nomenclature[A]{{PNG}}{Portable Network Graphics}
\nomenclature[A]{{SGD}}{Saccharomyces Genome Database}
\nomenclature[A]{{CPU}}{Central Processing Unit}
%\nomenclature[N]{VGG16}{A Convolutional Neural Network Model}
\nomenclature[N]{YouTubeCrash}{Dataset collected from dashcam videos uploaded on YouTube}
\nomenclature[N]{GTACrash}{Dataset collected from a video game named
Grand Theft Auto 5}

\printnomenclature[2cm]

\end{document}